\documentclass{new_tlp}

\usepackage[T1]{fontenc}
\usepackage{amsmath}
\usepackage{url}
\usepackage[all]{foreign}
\usepackage{tikz}
\usepackage{graphics}
\usepackage[caption=false]{subfig}

\newcommand{\tb}{\textbf}
\newcommand{\ti}{\textit}
\newcommand{\ttt}{\texttt}
\newcommand{\ts}{\textsf}
\newcommand{\mt}{\mathtt}
\newcommand{\ms}{\mathsf}
\newcommand{\mc}{\mathcal}

\newcommand{\B}[1]{\ensuremath{\tb{B}_\ms{#1}}}
\newcommand{\E}[1]{\ensuremath{\tb{E}_\ms{#1}}}
\newcommand{\C}[1]{\ensuremath{\tb{C}_\ms{#1}}}

\newcommand{\mal}{\ensuremath{m\mc{A}^*}}
\newcommand{\mar}{\ensuremath{m\mc{A}^\rho}}
\newcommand{\init}{\textnormal{\tb{initially}}\ }

\newcommand*{\axT}{\textbf{T}}

\submitted{5 June 2020}
\revised{23 June 2020}
\accepted{30 June 2020}

\newtheorem{definition}{Definition}

\newtheorem{proposition}{Proposition}

\title{Modelling Multi-Agent Epistemic Planning in ASP}

\author[A. Burigana et al.]{
	Alessandro Burigana, Francesco Fabiano, Agostino Dovier\\%
	University of Udine, Udine, Italy\\%
	\email{burigana.alessandro@spes.uniud.it}\\%
	\email{\{francesco.fabiano,agostino.dovier\}@uniud.it}%
	\and
	Enrico Pontelli\\%
	New Mexico State University, Las Cruces, NM, USA\\%
	\email{epontell@cs.nmsu.edu}%
}
 
\begin{document}
	\maketitle%


	\begin{abstract}\label{sec:abstract}
		Designing agents that reason and act upon the world has always been one of the main objectives of the Artificial Intelligence community. While for planning in ``simple'' domains the agents can solely rely on facts about the world, in several contexts, \textit{e.g.}, economy, security, justice and politics, the mere knowledge of the world could be insufficient to reach a desired goal. In these scenarios, \textit{epistemic} reasoning, \textit{i.e.}, reasoning about agents' beliefs about themselves and about other agents' beliefs, is essential to design winning strategies.
		This paper addresses the problem of reasoning in multi-agent epistemic settings exploiting declarative programming techniques. In particular, the paper presents an actual implementation of a multi-shot \textit{Answer Set Programming}-based planner that can reason in multi-agent epistemic settings, called \textsf{PLATO} (e\textbf{P}istemic mu\textbf{L}ti-agent \textbf{A}nswer se\textbf{T} programming s\textbf{O}lver). The ASP paradigm enables a concise and elegant design of the planner, w.r.t. other imperative implementations, facilitating the development of formal verification of correctness.
		The paper shows how the planner, exploiting an ad-hoc epistemic state representation and the efficiency of ASP solvers, has competitive performance results on benchmarks collected from the literature. It is under consideration for acceptance in TPLP.%
	\end{abstract}%



	\section{Introduction}\label{sec:motivation}
		The research area of \emph{Reasoning about Actions and Change (RAC)} has been particularly active in recent years, motivated by the wider introduction of autonomous systems and the use of multi-agent techniques in a variety of domains (e.g., cyber-physical systems). The role of logic programming has been central to RAC research, especially thanks to the use of logic programming to formalize the semantics of high level action languages and to experiment with different extensions of such languages~\cite{DBLP:journals/jlp/GelfondL93}.
	
		Over the years, several action languages (\eg $\cal A$, $\cal B$, and $\cal C$) have been developed, as discussed by Gelfond et al.~\cite{gelfond1998action}. Each of these languages addresses important problems in RAC. Action languages have also provided the foundations for several successful approaches to automated planning~\cite{CastelliniGT01,DBLP:conf/lpnmr/SonTGM05}. 
		
		In the special case of \emph{multi-agent domains}, an agent's action may not just change the world (and possibly the agent's own knowledge), but also may change other agents' knowledge and beliefs. Similarly, the goals of an agent in a multi-agent domain, may involve not only reaching a desirable configuration of the world, but may also involve affecting the knowledge and beliefs of other agents about the world. While the literature about planning in multi-agent domains is rich~\cite{DBLP:journals/aim/Durfee99a,DBLP:journals/amai/WeerdtBTW03,GoldmanZ04,DBLP:journals/mags/WeerdtC09,DBLP:journals/tplp/DovierFP13}, relatively fewer efforts have explored the challenges of planning in multi-agent domains in presence of goals and actions that rely on and manipulate agents' knowledge and beliefs.
		In previous work~\cite{Bar15}, we proposed a high level action language, $ \mal $, providing such features as: (i) actions that can change the world; (ii) actions that can impact either the knowledge of the agent or the beliefs and knowledge of other agents; (iii) actions that can affect agents' awareness of other events' occurrence. $ \mal $ has received an updated semantics, based on \emph{possibilities}~\cite{Fab19,Fab20}, which offered several advantages, in terms of simplicity and compactness of state representations. Two different planners have been proposed by Le et al. and Fabinano et al.~\cite{Le18,Fab20}, demonstrating the feasibility of planning in the domains described by $ \mal $.
		
		In this paper, as Baral et al.~\cite{Bar10}, we explore the use of logic programming, in the form of Answer Set Programming (ASP), to provide a novel implementation of a multi-agent epistemic planner. The implementation supports planning domains described using our possibilities-based multi-agent action language. The interest in this research direction derives from the desire of having a planner which is usable, efficient, and yet encoded using a declarative language. The declarative encoding allows us to provide formal proofs of correctness, which are presented in this paper. The declarative encoding will furnish a framework to explore a diversity of aspects of multi-agent epistemic planning, such as the impact of different optimizations (e.g., heuristics, avoidance of repeated states), the use of different semantics, and the introduction of extensions of the original action language.
		
		The implementation relies on the multi-shot capabilities of the \ti{clingo} solver. The encodings, a discussion about the nature of possibilities (Definition~\ref{def:poss}) and complete proofs of Propositions~\ref{prop:entail}--\ref{prop:ontic_corr} are available at \url{http://clp.dimi.uniud.it/sw/}.


	\section{Multi-Agent Epistemic Planning}\label{sec:mep}
		Let us begin by introducing the core elements of \emph{Multi-agents Epistemic Planning (MEP).} Let $ \mc{AG} $ be a finite set of agents and $ \mc{F} $ be a finite set of propositional variables, called \ti{fluents}. Fluents allow us to describe the properties of the world in which the agents operate. A \ti{possible world} is a representation of a possible configuration of the world, and it is described by a subset of $ \mc{F}$ (intuitively, those fluents that are \ti{true} in that world). Agents often have incomplete knowledge of the world, thus requiring the agent to deal with a set of possible worlds; the incomplete knowledge applies also to each agent's knowledge/beliefs about other agent's knowledge/beliefs.
		In epistemic planning, each action can be performed by an agent $ \ts{ag} \in \mc{AG}$. The effect of an action can either change the physical state of the world (\ie the fluents) or the agents' beliefs. Specifically, we want to deal with the agents' beliefs about both the world and the other agents' beliefs. To this end we make use of a logic that is concerned with information change, namely \ti{Dynamic Epistemic Logic (DEL)}~\cite{van2007}.
		
		We first introduce its syntax. A \ti{fluent formula} is a propositional formula built using the fluents in $ \mc{F}$. Fluent formulae allow us to express properties about a single possible world. To assert properties about what an agent believes, we use the modal operator $ \B{i} $, where $ \ts{i} \in \mc{AG}$. We read a formula $ \B{i} \varphi $ as ``agent \ts{i} believes $ \varphi $''. Given a nonempty set of agents $\alpha \subseteq \mc{AG}$, the \ti{group operators} $ \E{\alpha} $ and $ \C{\alpha}$, that intuitively represent the belief and the common belief of $\alpha$, respectively, will be also used.
	
		\begin{definition}[Belief formula]
			A \ti{belief formula} is defined recursively as follows:
			\begin{itemize}
				\item A fluent formula is a belief formula.
				\item If $ \varphi_1, \varphi_2 $ are belief formulae, then $ \neg \varphi_1 $, $ \varphi_1 \vee \varphi_2 $ and $ \varphi_1 \wedge \varphi_2 $ are belief formulae.
				\item If $ \varphi $ is a belief formula and $ \ms{ag} \in \mc{AG} $ is an agent, then $ \B{ag} \varphi $ is a belief formula.
				\item If $ \varphi $ is a belief formula and $ \emptyset \neq \alpha \subseteq \mc{AG} $, then $ \E{\alpha} \varphi $ and $ \C{\alpha} \varphi $ are belief formulae.
			\end{itemize}
		\end{definition}
			
		\noindent The semantics of DEL formulae is traditionally expressed using \ti{pointed Kripke structures}~\cite{Kri63}; in previous work~\cite{Fab19,Fab20}, we provided a semantics based on the concept of \ti{possibilities}. 
			
		\begin{definition}[Possibility~\cite{Ger97}]\label{def:poss}
			\begin{itemize}
				\item A \ti{possibility} \ts{u} is a function that assigns to each fluent $ \ms{f} \in \mc{F} $ a truth value $ \ms{u}(\ms{f}) \in \{0, 1\} $ and to each agent $ \ms{ag} \in \mc{AG} $ an information state $ \ms{u}(\ms{ag}) = \sigma $;
				\item An \ti{information state} $ \sigma $ is a set of possibilities.
			\end{itemize}
		\end{definition}
		
		\noindent Possibilities allow us to capture the concept of \ti{epistemic state} (briefly, \ti{e-state}). E-states consist of two components: information about the possible worlds and information about the agents' beliefs. Let \ts{u} be a possibility. The assignment of truth values $ \ms{u}(\ms{f}) $ for each fluent $ \ms{f} \in \mc{F} $ encodes a possible world; the assignment of information states to an agent $ \ms{ag} \in \mc{AG} $ captures the beliefs of \ts{ag}. Information states encode the same information represented by the edges of a Kripke structure; that is, an information state $ \ms{u}(\ms{ag}) $ is comparable to the set of worlds reached by \ts{ag} from the world \ts{u} in a Kripke structure.
		If $ \ms{u}(\ms{f}) = 0 $, then, in the possible world represented by \ts{u}, the fluent \ts{f} is false. Similarly, if $ \ms{u}(\ms{ag}) = \{\ts{v}\} $, then (in the possibility \ts{u}) the agent \ts{ag} believes only the possibility \ts{v}. Since possibilities are non-well-founded objects (\ie we do \ti{not} require the sets to be well-founded), the concepts of \ti{state} and \ti{possible world} collapse. In fact, a possibility contains both the information of a possible world (the interpretation of the fluents) and the information about the agents' beliefs (represented by other possibilities). Hence, we denote the state/possible world that represents the \ti{real world} as the \ti{pointed possibility}. Due to space constraints, we refer the interested reader to the supplementary documents (available at \url{http://clp.dimi.uniud.it/sw/}) and to Gerbrandy et al. and Fabiano et al.~\cite{Ger97,Fab19} for a more complete discussion on the nature of possibilities.
	
		\begin{definition}[Entailment w.r.t. possibilities~\cite{Fab19}]
			Let the belief formulae $ \varphi, \varphi_1, \varphi_2 $, a fluent \ts{f}, an agent \ts{ag}, a (non-empty) group of agents $ \alpha $, and a possibility \ts{u} be given.
			\begin{enumerate}
			\renewcommand{\theenumi}{\roman{enumi}}
				\item $ \ms{u} \models \ms{f} $ if $ \ms{u}(\ms{f}) = 1 $;
				\item $ \ms{u} \models \neg \varphi $ if $ \ms{u} \not\models \varphi $;
				\item $ \ms{u} \models \varphi_1 \vee \varphi_2 $ if $ \ms{u} \models \varphi_1 $ or $ \ms{u}\models \varphi_2 $;
				\item $ \ms{u} \models \varphi_1 \wedge \varphi_2 $ if $ \ms{u} \models \varphi_1 $ and $ \ms{u}\models \varphi_2 $;
				\item $ \ms{u} \models \B{ag} \varphi $ if for each $ \ms{v} \in \ms{u}(\ms{ag}) $ it holds that $ \ms{v} \models \varphi $;
				\item $ \ms{u} \models \E{\alpha} \varphi $ if for all $ \ms{ag} \in \alpha $ it holds that $ \ms{u} \models \B{ag} \varphi $ ;
				\item $ \ms{u} \models \C{\alpha} \varphi $ if $ \ms{u} \models \E{\alpha}^k \varphi $ for every $ k \geq 0 $, where $ \E{\alpha}^0 \varphi = \varphi$ and $ \E{\alpha}^{k+1} \varphi = \E{\alpha}(\E{\alpha}^k \varphi) $.
			\end{enumerate}
		\end{definition}
		
		\noindent We say that an agent \ti{believes} a belief formula $ \varphi $ w.r.t. a given possibility if all of the possibilities within its information state entail $ \varphi $. Common belief requires all agents in $\alpha $ to believe $\varphi$, that \ti{all the agents in $\alpha$ believe $\varphi$} and so on \ti{ad infinitum}.
		
		\begin{definition}[MEP domain]\label{def:MEPdomain}
			A \ti{multi-agent epistemic planning domain} is a tuple $ D = \langle \mc{F}, \mc{AG}, \mc{A}, \varphi_i, \varphi_g \rangle $, where: 
			\begin{enumerate}
			\renewcommand{\theenumi}{\roman{enumi}}
				\item $ \mc{F} $ is the finite set of \ti{fluents} of $ D $;
				\item $ \mc{AG} $ is the finite set of \ti{agents} of $ D $;
				\item $ \mc{A} $ represents the set of \ti{actions} of $ D $;
				\item $ \varphi_i $ is the belief formula that describes the \ti{initial conditions}; and
				\item $ \varphi_g $ is the belief formula that describes the \ti{goal conditions} that we want to achieve.
			\end{enumerate}
		\end{definition}
		
		\noindent A domain contains the information needed to describe a planning problem in a multi-agent epistemic setting. Given a domain $ D $ we refer to its elements through the parenthesis operator; \eg the fluent set of $ D $ will be denoted by $ D(\mc{F}) $. An \ti{action instance} $ \ms{a}\langle \ms{ag} \rangle \in D(\mc{AI}) = D(\mc{A}) \times D(\mc{AG}) $ identifies the execution of action $ \ms{a} $ by agent $ \ms{ag} $. Let $ D(\mc{S}) $ be the set of states reachable from $ D(\varphi_i) $ with a finite sequence of actions. The \ti{transition function} $ \Phi : D(\mc{AI}) \times D(\mc{S}) \rightarrow D(\mc{S})\ \cup\ \{\emptyset\} $ allows us to formalize the semantics of action instances (the result is the empty set if the action instance is not executable).
		
\begin{figure}
			\centering%
			\subfloat[Picture of \ts{w}.]{\scalebox{0.55}{
				\begin{tikzpicture}[x=0.75pt,y=0.75pt,yscale=-1,xscale=1]
					\draw    (157.85,100.75) -- (436.3,164.8) ;
					\draw [shift={(438.25,165.25)}, rotate = 192.95] [fill={rgb, 255:red, 0; green, 0; blue, 0 }  ][line width=0.75]  [draw opacity=0] (10.72,-5.15) -- (0,0) -- (10.72,5.15) -- (7.12,0) -- cycle    ;
	
					\draw    (438.25,100.75) -- (159.8,164.8) ;
					\draw [shift={(157.85,165.25)}, rotate = 347.05] [fill={rgb, 255:red, 0; green, 0; blue, 0 }  ][line width=0.75]  [draw opacity=0] (10.72,-5.15) -- (0,0) -- (10.72,5.15) -- (7.12,0) -- cycle    ;
	
					\draw    (145.3,101.6) -- (145.3,164.6) ;
					\draw [shift={(145.3,166.6)}, rotate = 270] [fill={rgb, 255:red, 0; green, 0; blue, 0 }  ][line width=0.75]  [draw opacity=0] (10.72,-5.15) -- (0,0) -- (10.72,5.15) -- (7.12,0) -- cycle    ;
	
					\draw    (451,102.6) -- (451,162.6) ;
					\draw [shift={(451,164.6)}, rotate = 270] [fill={rgb, 255:red, 0; green, 0; blue, 0 }  ][line width=0.75]  [draw opacity=0] (10.72,-5.15) -- (0,0) -- (10.72,5.15) -- (7.12,0) -- cycle    ;
	
					\draw    (160.3,88) -- (434.3,88) ;
					\draw [shift={(436.3,88)}, rotate = 180] [fill={rgb, 255:red, 0; green, 0; blue, 0 }  ][line width=0.75]  [draw opacity=0] (10.72,-5.15) -- (0,0) -- (10.72,5.15) -- (7.12,0) -- cycle    ;
					\draw [shift={(158.3,88)}, rotate = 0] [fill={rgb, 255:red, 0; green, 0; blue, 0 }  ][line width=0.75]  [draw opacity=0] (10.72,-5.15) -- (0,0) -- (10.72,5.15) -- (7.12,0) -- cycle    ;
	
					\draw    (158.3,178) -- (434.3,178) ;
					\draw [shift={(436.3,178)}, rotate = 180] [fill={rgb, 255:red, 0; green, 0; blue, 0 }  ][line width=0.75]  [draw opacity=0] (10.72,-5.15) -- (0,0) -- (10.72,5.15) -- (7.12,0) -- cycle    ;
					\draw [shift={(156.3,178)}, rotate = 0] [fill={rgb, 255:red, 0; green, 0; blue, 0 }  ][line width=0.75]  [draw opacity=0] (10.72,-5.15) -- (0,0) -- (10.72,5.15) -- (7.12,0) -- cycle    ;
	
					\draw    (131.75,92.5) .. controls (95.12,70.72) and (138.15,31.13) .. (148.98,73.78) ;
					\draw [shift={(149.3,75.1)}, rotate = 257.08] [fill={rgb, 255:red, 0; green, 0; blue, 0 }  ][line width=0.75]  [draw opacity=0] (10.72,-5.15) -- (0,0) -- (10.72,5.15) -- (7.12,0) -- cycle    ;
	
					\draw    (465.42,92.5) .. controls (493.02,62.9) and (457.03,31.24) .. (446.71,74.37) ;
					\draw [shift={(446.41,75.7)}, rotate = 282.37] [fill={rgb, 255:red, 0; green, 0; blue, 0 }  ][line width=0.75]  [draw opacity=0] (10.72,-5.15) -- (0,0) -- (10.72,5.15) -- (7.12,0) -- cycle    ;
	
					\draw    (464.29,174) .. controls (494.83,201.19) and (458.42,225.85) .. (447.22,192.92) ;
					\draw [shift={(446.72,191.38)}, rotate = 433.28999999999996] [fill={rgb, 255:red, 0; green, 0; blue, 0 }  ][line width=0.75]  [draw opacity=0] (10.72,-5.15) -- (0,0) -- (10.72,5.15) -- (7.12,0) -- cycle    ;
	
					\draw    (131.12,173.65) .. controls (98.92,194.9) and (135.57,235.08) .. (149.7,191.81) ;
					\draw [shift={(150.12,190.47)}, rotate = 466.9] [fill={rgb, 255:red, 0; green, 0; blue, 0 }  ][line width=0.75]  [draw opacity=0] (10.72,-5.15) -- (0,0) -- (10.72,5.15) -- (7.12,0) -- cycle    ;
	
					\draw (145.1,89) node [scale=2.5] [align=left] {\ts{w}};
					\draw (453,85) node [scale=2] [align=left] {\ts{w'}};
					\draw (145.1,178) node [scale=2] [align=left] {\ts{v}};
					\draw (455,176) node [scale=2] [align=left] {\ts{v'}};
	
					\draw  [color={rgb, 255:red, 255; green, 255; blue, 255 }  ,draw opacity=0.2 ][fill={rgb, 255:red, 255; green, 255; blue, 255 }  ,fill opacity=1 ]  (103.69,68.36) .. controls (103.69,63.39) and (110.63,59.36) .. (119.19,59.36) .. controls (127.75,59.36) and (134.69,63.39) .. (134.69,68.36) .. controls (134.69,73.33) and (127.75,77.36) .. (119.19,77.36) .. controls (110.63,77.36) and (103.69,73.33) .. (103.69,68.36) -- cycle  ;
					\draw (119.19,68.36) node [scale=1] [align=left] {\ts{A}, \ts{B}};
	
					\draw  [color={rgb, 255:red, 255; green, 255; blue, 255 }  ,draw opacity=0.2 ][fill={rgb, 255:red, 255; green, 255; blue, 255 }  ,fill opacity=1 ]  (92.19,190.36) .. controls (92.19,185.39) and (102.49,181.36) .. (115.19,181.36) .. controls (127.9,181.36) and (138.19,185.39) .. (138.19,190.36) .. controls (138.19,195.33) and (127.9,199.36) .. (115.19,199.36) .. controls (102.49,199.36) and (92.19,195.33) .. (92.19,190.36) -- cycle  ;
					\draw (115.19,190.36) node [scale=1] [align=left] {\ts{A}, \ts{B}, \ts{C}};
	
					\draw  [color={rgb, 255:red, 255; green, 255; blue, 255 }  ,draw opacity=0.2 ][fill={rgb, 255:red, 255; green, 255; blue, 255 }  ,fill opacity=1 ]  (455.19,190.36) .. controls (455.19,185.39) and (465.49,181.36) .. (478.19,181.36) .. controls (490.9,181.36) and (501.19,185.39) .. (501.19,190.36) .. controls (501.19,195.33) and (490.9,199.36) .. (478.19,199.36) .. controls (465.49,199.36) and (455.19,195.33) .. (455.19,190.36) -- cycle  ;
					\draw (478.19,190.36) node [scale=1] [align=left] {\ts{A}, \ts{B}, \ts{C}};
	
					\draw  [color={rgb, 255:red, 255; green, 255; blue, 255 }  ,draw opacity=0.2 ][fill={rgb, 255:red, 255; green, 255; blue, 255 }  ,fill opacity=1 ]  (457.69,68.36) .. controls (457.69,63.39) and (464.63,59.36) .. (473.19,59.36) .. controls (481.75,59.36) and (488.69,63.39) .. (488.69,68.36) .. controls (488.69,73.33) and (481.75,77.36) .. (473.19,77.36) .. controls (464.63,77.36) and (457.69,73.33) .. (457.69,68.36) -- cycle  ;
					\draw (473.19,68.36) node [scale=1] [align=left] {\ts{A}, \ts{B}};
	
					\draw  [color={rgb, 255:red, 255; green, 255; blue, 255 }  ,draw opacity=0.2 ][fill={rgb, 255:red, 255; green, 255; blue, 255 }  ,fill opacity=1 ]  (275.05,178.36) .. controls (275.05,173.39) and (285.35,169.36) .. (298.05,169.36) .. controls (310.75,169.36) and (321.05,173.39) .. (321.05,178.36) .. controls (321.05,183.33) and (310.75,187.36) .. (298.05,187.36) .. controls (285.35,187.36) and (275.05,183.33) .. (275.05,178.36) -- cycle  ;
					\draw (298.05,178.36) node [scale=1] [align=left] {\ts{A}, \ts{B}, \ts{C}};
	
					\draw  [color={rgb, 255:red, 255; green, 255; blue, 255 }  ,draw opacity=0.2 ][fill={rgb, 255:red, 255; green, 255; blue, 255 }  ,fill opacity=1 ]  (281.8,88) .. controls (281.8,83.03) and (288.74,79) .. (297.3,79) .. controls (305.86,79) and (312.8,83.03) .. (312.8,88) .. controls (312.8,92.97) and (305.86,97) .. (297.3,97) .. controls (288.74,97) and (281.8,92.97) .. (281.8,88) -- cycle  ;
					\draw (297.3,88) node [scale=1] [align=left] {\ts{A}, \ts{B}};
	
					\draw  [color={rgb, 255:red, 255; green, 255; blue, 255 }  ,draw opacity=0.2 ][fill={rgb, 255:red, 255; green, 255; blue, 255 }  ,fill opacity=1 ]  (137.8,133) .. controls (137.8,128.03) and (141.61,124) .. (146.3,124) .. controls (150.99,124) and (154.8,128.03) .. (154.8,133) .. controls (154.8,137.97) and (150.99,142) .. (146.3,142) .. controls (141.61,142) and (137.8,137.97) .. (137.8,133) -- cycle  ;
					\draw (146.3,133) node [scale=1] [align=left] {\ts{C}};
	
					\draw  [color={rgb, 255:red, 255; green, 255; blue, 255 }  ,draw opacity=0.2 ][fill={rgb, 255:red, 255; green, 255; blue, 255 }  ,fill opacity=1 ]  (441.5,133) .. controls (441.5,128.03) and (445.31,124) .. (450,124) .. controls (454.69,124) and (458.5,128.03) .. (458.5,133) .. controls (458.5,137.97) and (454.69,142) .. (450,142) .. controls (445.31,142) and (441.5,137.97) .. (441.5,133) -- cycle  ;
					\draw (450,133) node [scale=1] [align=left] {\ts{C}};
	
					\draw  [color={rgb, 255:red, 255; green, 255; blue, 255 }  ,draw opacity=0.2 ][fill={rgb, 255:red, 255; green, 255; blue, 255 }  ,fill opacity=1 ]  (220.8,116) .. controls (220.8,111.03) and (224.61,107) .. (229.3,107) .. controls (233.99,107) and (237.8,111.03) .. (237.8,116) .. controls (237.8,120.97) and (233.99,125) .. (229.3,125) .. controls (224.61,125) and (220.8,120.97) .. (220.8,116) -- cycle  ;
					\draw (229.3,116) node [scale=1] [align=left] {\ts{C}};
	
					\draw  [color={rgb, 255:red, 255; green, 255; blue, 255 }  ,draw opacity=0.2 ][fill={rgb, 255:red, 255; green, 255; blue, 255 }  ,fill opacity=1 ]  (367.8,116) .. controls (367.8,111.03) and (371.61,107) .. (376.3,107) .. controls (380.99,107) and (384.8,111.03) .. (384.8,116) .. controls (384.8,120.97) and (380.99,125) .. (376.3,125) .. controls (371.61,125) and (367.8,120.97) .. (367.8,116) -- cycle  ;
					\draw (376.3,116) node [scale=1] [align=left] {\ts{C}};
				\end{tikzpicture}
			}\label{subfig-state_as_pos:2}}
			\hspace{10pt}
			\subfloat[System of equations of \ts{w}.]{\scalebox{0.85}{\raisebox{5em}{
				$
					\begin{cases}
						\ms{w}        & = \{
							(\ms{A},\{\ms{w}, \ms{w'}\}),
							(\ms{B},\{\ms{w}, \ms{w'}\}),
							(\ms{C},\{\ms{v}, \ms{v'}\}),
							\ms{f},
							\ms{g},
							\ms{h}%
						\}                         \\
						\ms{w'} & =\{
							(\ms{A},\{\ms{w}, \ms{w'}\}),
							(\ms{B},\{\ms{w}, \ms{w'}\}),
							(\ms{C},\{\ms{v}, \ms{v'}\}),
							\ms{g},
							\ms{h}%
						\}                         \\
						\ms{v}        & = \{
							(\ms{A},\{\ms{v}, \ms{v'}\}),
							(\ms{B},\{\ms{v}, \ms{v'}\}),
							(\ms{C},\{\ms{v}, \ms{v'}\}),
							\ms{f},
							\ms{h}%
						\}                         \\
						\ms{v'} & = \{
							(\ms{A},\{\ms{v}, \ms{v'}\}),
							(\ms{B},\{\ms{v}, \ms{v'}\}),
							(\ms{C},\{\ms{v}, \ms{v'}\}),
							\ms{h}%
						\}                         \\
					\end{cases}
				$
			}}\label{subfig-state_as_pos:1}}%
			\caption{Two equivalent representations of a generic possibility $ \ms{w} $. The possibility is expanded for clarity. Only ``true'' fluents are put in the set (rather than all pairs $(\ms{f},1), (\ms{g},1), (\ms{h},1) $). The interpretation of the fluents is the same in both figures.}
			\label{fig:state_as_pos}
		\end{figure}
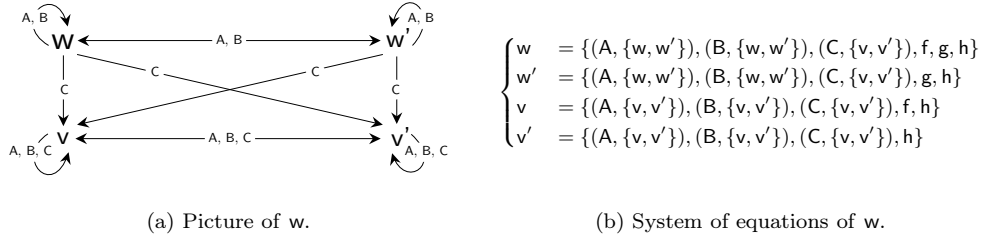

		Possibilities are objects with a non-well-founded nature~\cite{Acz88}. This allows us to represent them by means of both a \ti{picture} (\ie a \ti{pointed graph}) and a \ti{system of equations}, which are the standard representations for non-well-founded sets. In Figure~\ref{fig:state_as_pos} an example of a generic possibility illustrated using these two representations.
		
		\begin{definition}[Decoration and picture~\cite{Acz88}]
			A \ti{decoration} of a graph $ G = (V, E) $ is a function $ \delta $ that assigns to each node $ \ms{n} \in V $ a (non-well-founded) set $ \delta_{\ms{n}}$, whose elements are the sets assigned to the successors of $\ms{n}$ in the graph, \ie $ \delta_{\ms{n}} = \{\delta_{\ms{n'}} : (n, n') \in E \} $.
			Given a \ti{pointed} graph $ (G, \ms{n}) $ (\ie a graph with a node $ \ms{n} \in V $ identified), if $ \delta $ is a decoration of $ G $, then $ (G, \ms{n}) $ is a \ti{picture} of the set $ \delta_\ms{n} $.
		\end{definition}
		
		\noindent Pictures also allow us to use graph terminology (edges, labels, reachability, etc.) when referring to possibilities. Given a possibility \ts{u} and its picture $ (G_\ms{u}, \ms{n_u}) $, when we refer to a ``(labeled) edge'' of \ts{u}, we actually allude to its picture. When the context is clear we will use such terminology to refer directly to a possibility.
		
		The non-well-founded nature of possibilities allow us to characterize the state equality through \ti{bisimulation} (see Dovier~\cite{DBLP:conf/iclp/Dovier15} for a brief introduction). In fact, two decorations with bisimilar labeled graphs are represented by the same possibility.
		
		\paragraph{\bf Knowledge or belief.}
			As pointed out in the previous paragraphs the modal operator $\B{ag}$ represents the worlds' relations in an e-state. As expected, different relations' properties imply different meanings for $\B{ag}$. In particular, we are interested in representing the agents' knowledge or beliefs. The accepted formalization for such concepts relies on the \tb{S5} and \tb{KD45} axioms, respectively. In fact, when a \ti{relation}---represented by the \ti{edges} in a Kripke structure and by the \emph{information state} in a possibility---respects all the axioms presented in Table~\ref{tab:axioms}, it is called an \tb{S5}-relation and it encodes the concept of knowledge; similarly, when the relation encodes all such axioms but \axT, we obtain a \tb{KD45}-relation, that characterizes the concept of belief. Following these characterization we will refer to the logics of knowledge and belief as \tb{S5} and \tb{KD45} logic, respectively.
			\begin{table}[h]
				\centering
				\includegraphics{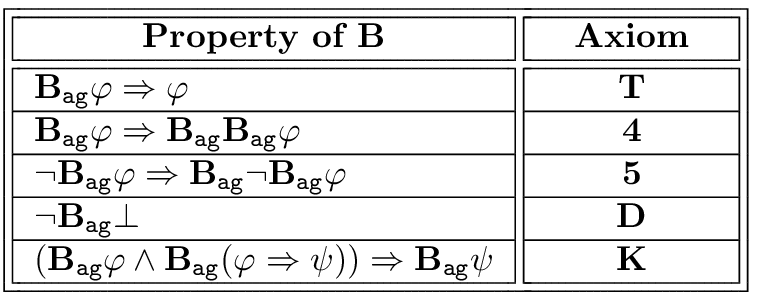}
				\caption{Knowledge and beliefs axioms.}
				\label{tab:axioms}
			\end{table}
				
			\noindent Intuitively the difference between the two logics is that an agent cannot \textit{know} something that is not true in \tb{S5}, but she can \textit{believe} it in \tb{KD45}. Our planner deals with e-states that comply with the axioms of \tb{KD45}. However, it is possible to encode a domain in such a way that, when an action is performed, the resulting e-state is consistent with the axiom \tb{T}. In this way we are able to reason within the \tb{S5} logic. As explained in the following pages, we only require the initial state to satisfy all the \tb{S5} axioms. As this introduction is not supposed to explore in depth this topic, we will not go into further detail and we address the interested reader to Fagin et al.~\cite{fagin2004reasoning}.
	
		\paragraph{\bf The language $\boldsymbol{\mar}$.}\label{sec:efp}
			The planner EFP 2.0, introduced by Fabiano et al. in previous work~\cite{Fab20}, is able to reason on epistemic domains. Domain instances are encoded using the action language $\mar$~\cite{Fab19}, in turn inspired by the language $ \mal $~\cite{Bar15}. The main difference between such languages lies in their semantics: while $\mal$ is based on Kripke structures and \ti{updated models}~\cite{Bar15}, $\mar$ is given in terms of possibilities~\cite{Fab20}.
			
			The languages $ \mal $ and $ \mar $ both allow three different types of action: i) \ti{ontic} (or \ti{world-altering}) actions that are used to change the properties of the world (\ie the truth value of fluents); ii) \ti{sensing} and iii) \ti{announcement} actions that are performed by an agent to change her beliefs about the world and to affect other agents' beliefs, respectively.
	
			The action languages also allow to specify, for each action instance \ts{a}, the \ti{observability relation} of each agent. Namely, an agent may be \ti{fully observant}, \ti{partially observant}, or \ti{oblivious} w.r.t. \ts{a}. If an agent is fully observant, then she is aware of both the execution of the action instance and its effects; she is partially observant if she is only aware of the action execution but not of the outcomes; she is oblivious if she is ignorant of the execution of the action.
	
			Given a domain $ D $, an action instance $ \ms{a} \in D(\mc{AI}) $, a fluent literal $ \ms{f} $, a fluent formula $ \phi $ and the belief formula $ \varphi $, we introduce the syntax of $ \mar $ as follows:
			\begin{itemize}
				\item $ \text{\tb{executable} } \ms{a} \text{ \tb{if} } \varphi $: captures the \ti{executability conditions};
				\item $ \ms{a} \text{ \tb{causes} } \ms{f} \text{ \tb{if} } \varphi$: captures the effects of \ti{ontic} actions;
				\item $ \ms{a} \text{ \tb{determines} } \ms{f} \text{ \tb{if} } \varphi$: captures the effects of \ti{sensing} actions;
				\item $ \ms{a} \text{ \tb{announces} } \phi \text{ \tb{if} } \varphi$: captures the effects of \ti{announcement} actions;
				\item $ \ms{ag} \text{ \tb{observes} } \ms{a} \text{ \tb{if} } \varphi$: captures \ti{fully observant} agents for an action; and
				\item $ \ms{ag} \text{ \tb{aware\_of} } \ms{a} \text{ \tb{if} } \varphi$: captures \ti{partially observant} agents for a given action.
			\end{itemize}
				
			\noindent Notice that if we do not state otherwise, an agent will be considered oblivious. Finally, statements of the form $ \text{\tb{initially} } \varphi $ and $ \text{\tb{goal} } \varphi $ capture the initial and goal conditions, respectively. The formulae $ \varphi_i $ and $ \varphi_g $ of a domain are obtained by a conjunction of the initial conditions and of the goal conditions, respectively.
	
		\paragraph{\bf Finitary S5 theories.}
			Given a generic belief formula $ \varphi $ it is possible to generate infinitely many (initial) states that satisfy $ \varphi $ (see Son et al.~\cite{Son14} for a complete introduction). To overcome this problem, we use the following notion and result.
	
			\begin{definition}[Finitary \tb{S5}-theory~\cite{Son14}]\label{def:s5_theory}
				Let $ \phi $ be a fluent formula and let $ \ms{i} \in \mc{AG} $ be an agent. A \ti{finitary \tb{S5}-theory} is a collection of formulae of the form (we use the short form $ \C{\ } \phi $ instead of $ \C{\mc{AG}} \phi $):
				\begin{center}
					\hfill
					$   (i)\ \phi \hfill $
					$  (ii)\ \C{\ } \phi \hfill $
					$ (iii)\ \C{\ } (\B{i} \phi \vee \B{i} \neg \phi) \hfill $
					$  (iv)\ \C{\ } (\neg \B{i} \phi \wedge \neg \B{i} \neg \phi) \hfill $
				\end{center}
				
				\noindent Moreover, we require each fluent $ \ms{f} \in \mc{F} $ to appear in at least one of the formulae $(ii)$--$(iv)$ (for at least one agent $ \ms{i} \in \mc{AG} $).
			\end{definition}
				
			\noindent As shown by Son et al., a finitary \tb{S5}-theory has finitely many S5-models up to equivalence (\ie bisimulation). We therefore require that the set of formulae $ \{\varphi \mid [\text{\tb{initially} } \varphi] \in D \} $ is a \ti{finitary \tb{S5}-theory}. Moreover, in Section~\ref{par:optimizations} we explain how the generation of a unique initial state is achieved. It is important to notice that this requirement applies only when the initial state description is given by means of a set of belief formulae. On the other hand, whenever the initial state is explicitly described, we do not impose any limitation. This allows us to simplify the initial state generation w.r.t. some other approaches~\cite{van2002tractable,bolander2011epistemic,lowe2011planning}, where the initial e-state is assumed to be explicitly described.
	

	\section{Multi-Shot Solving in ASP}\label{sec:asp}
		A general program $ P $ in the language ASP is a set of rules $ r $ of the form:
		$$ a_0 \leftarrow a_1, \dots, a_m, not\ a_{m+1}, \dots, not\ a_n $$
		
		\noindent where $ 0 \leq m \leq n $ and each element $ a_i $, with $0 \leq i \leq n$, is an \ti{atom} of the form $ p(t_1, \dots, t_k) $, $ p $ is a predicate symbol of arity $ k $ and $ t_1,\dots,t_k$ are terms built using variables, constants and function symbols. Negation-as-failure (naf) literals are of the form $ not\ a $, where $ a $ is an atom. Let $ r $ be a rule, we denote with $h(r)= a_0$ its \ti{head}, and $ B^+(r) = \{a_1, \dots, a_m\}$ and $ B^-(r)=\{a_{m+1}, \dots, a_n\} $ the positive and negative parts of its \ti{body}, respectively; we denote the body with $B(r) = \{a_1,\dots, not\ a_n\}$. A rule is called a \ti{fact} whenever $ B(r) = \emptyset $; a rule is a \ti{constraint} when its head is empty ($h(r)=\mathsf{false}$); if $ m = n $ the rule is a \ti{definite rule}. A \ti{definite program} consists of only definite rules.
		
		A term, atom, rule, or program is said to be \ti{ground} if it does not contain variables. Given a program $ P $, its \ti{ground instance} is
		the set of all ground rules obtained by substituting all variables in each rule with ground terms. In what follows we assume atoms, rules and programs to be \ti{grounded}. Let $ M $ be a set of ground atoms ($\mathsf{false} \notin M$) and let $ r $ be a rule: we say that $ M \models r$ if $ B^+(r) \not\subseteq M $ or $ B^-(r) \cap M \neq \emptyset $ or $ h(r) \in M $. $M$ is a \ti{model} of $P$ if $M\models r$ for each $r \in P$. The \ti{reduct} of a program $ P $ w.r.t.\ $ M $, denoted by $ P^M $, is the definite program obtained from $P$ as follows: (i) for each $ a \in M$, delete all the rules $r$ such that $a \in B^-(r)$, and (ii) remove all naf-literals in the the remaining rules. A set of atoms $ M $ is an \ti{answer set}~\cite{Gel88} of a program $ P $ if $M$ is the minimal model of $P^M$. A program $ P $ is \ti{consistent} if it admits an answer set.
		
		We will make use of the multi-shot declarations for ASP, i.e.\ statements of the form $ \text{\ttt{\#program} } sp(p_1, \dots, p_k) $,
		where $ sp $ is the name of the sub-program and the $ p_i$'s are its parameters~\cite{Ges17}. Precisely, if $ R $ is a \ti{list} of non-ground rules and declarations, with $R(sp)$ we denote the sub-program consisting of all the rules following the statement up to the next program declaration (or the end of the list). If the list does not start with a declaration, the default declaration \ttt{\#base} is implicitly added by \ti{clingo}. 

		An ASP program $ R $ is \ti{extensible} if it contains declarations of the form $ \text{\ttt{\#external} } a : B $, where $ a $ is an atom and $ B $ is a rule body. These declarations identify a set of atoms that are outside the scope of traditional ASP solving (e.g., they may not appear in the head of any rule). When we set $ a $ to true we can \ti{activate} all the rules $ r $ such that $ a \in B^+(r) $. Splitting the program allows us to control the grounding and solving phases of each sub-program by explicit instructions using a Python script.
	

	\section{Modeling Epistemic Multi-agent Planning using ASP}\label{sec:contribute}
		We present a multi-shot ASP encoding for solving a multi-agent epistemic planning domain $ D = \langle \mc{F}, \mc{AG}, \mc{A}, \varphi_i, \varphi_g \rangle $ (Definition~\ref{def:MEPdomain}) upon the possibilities based semantics described by Fabiano et al.~\cite{Fab20}. Its core elements are: the entailment of DEL formulae, the generation of the initial state and the transition function. The encoding implements a breadth-first search exploiting the multi-shot capabilities of \ti{clingo}.

		\paragraph{\bf Epistemic states.}\label{sec:prg_structure}
			As we discussed in Section~\ref{sec:mep}, the elements that we need to encode are the possible worlds and the agents' beliefs. We use atoms of the form \ttt{possible\_world(T, R, P)} and \ttt{believes(T1, R1, P1, T2, R2, P2, AG)}, respectively. Intuitively, the first atom identifies a possibility with a triple $ (\mt{T}, \mt{R}, \mt{P}) $, while the second encodes an ``edge'' between the possibilities $ (\mt{T1}, \mt{R1}, \mt{P1}) $ and $ (\mt{T2}, \mt{R2}, \mt{P2})$, labeled with the agent \ttt{AG}.
			
			Let us now focus in more detail on \ttt{possible\_world(T, R, P)}. \ttt{P} is the index of the possibility. The variables \ttt{T} and \ttt{R} represent the \ti{time} and the \ti{repetition} of the possibility \ttt{P}, respectively. It is important to notice that these two parameters are necessary to uniquely identify a possibility during the solving process. The first parameter tells us \ti{when} \ttt{P} was created: a possibility with time \ttt{T} is created after the execution of an action at time \ttt{T}. At a given time, it could be the case that two (or more) possibilities share both the values of \ttt{T} and \ttt{P}. Thus, a third value, the repetition \ttt{R}, is introduced with the only purpose to disambiguate between these cases. The update of repetitions will be explained during the analysis of the transition function.
			
			Intuitively, the index \ttt{P} is used during the generation of the initial state to name the initial possible worlds. Afterwards, when an action is performed, we create new possibilities by updating the values of \ttt{T} and \ttt{R}. We do not need to modify the value of \ttt{P} as well, since the update of time and repetition is designed to be univocal for each \ttt{P}.
		
			Let \ts{ag} be an agent and \ts{u} and \ts{v} be two possibilities represented by the triples $ (\mt{Tu}, \mt{Ru}, \mt{Pu}) $ and $ (\mt{Tv}, \mt{Rv}, \mt{Pv}) $, respectively. Then, we encode the fact that $ \ms{v} \in \ms{u}(\ms{ag}) $ with the atom \ttt{believes(Tu, Ru, Pu, Tv, Rv, Pv, ag)}. 
				
			The truth value of each fluent is captured by an atom of the form \ttt{holds(Tu, Ru, Pu, F)}. The truth of such atom captures the fact that $ \ms{u}(\mt{F}) = 1 $. Finally, we specify the pointed possibility, for a given time \ttt{T}, using atoms of the form \ttt{pointed(T, R, P)}. For readability purposes, in the following pages we will identify a possibility \ts{u} by \ttt{Pu} rather than by the triple $ (\mt{Tu}, \mt{Ru}, \mt{Pu}) $, when this will cause no ambiguity.

		\paragraph{\bf Entailment.}\label{sec:entailment}
			To verify if a given DEL formula \ttt{F} is entailed by a possibility, we use the predicate \ttt{entails(P, F)}, defined below (with some simplifications for readability).
			\begin{equation*}
				\begin{array}{llrl}
					\mt{entails     } & \mt{(P,}  & \mt{ F    )}      & \text{\ttt{:-}}\ \mt{     holds(P, F),\                       fluent(F).}         \vspace{3pt} \\
					\mt{entails     } & \mt{(P,}  & \mt{neg(F))     } & \text{\ttt{:-}}\ \mt{not\ entails(P, F ).}                                                 \\
					\mt{entails     } & \mt{(P,}  & \mt{and(F1, F2))} & \text{\ttt{:-}}\ \mt{     entails(P, F1),\ entails(P, F2).}                                \\
					\mt{entails     } & \mt{(P,}  & \mt{ or(F1, F2))} & \text{\ttt{:-}}\ \mt{     entails(P, F1).}                                                 \\
					\mt{entails     } & \mt{(P,}  & \mt{ or(F1, F2))} & \text{\ttt{:-}}\ \mt{                      entails(P, F2).}                   \vspace{3pt} \\
					\mt{not\_entails} & \mt{(P1,} & \mt{b(AG , F))}   & \text{\ttt{:-}}\ \mt{not\      entails(P2, F        ),\ believes(P1, P2, AG).}             \\
					\mt{entails     } & \mt{(P ,} & \mt{b(AG , F))}   & \text{\ttt{:-}}\ \mt{not\ not\_entails(P , b(AG , F)).}                       \vspace{3pt} \\
					\mt{not\_entails} & \mt{(P1,} & \mt{c(AGS, F))}   & \text{\ttt{:-}}\ \mt{not\      entails(P2, F        ),\ reaches(P1, P2, AGS).}              \\
					\mt{entails     } & \mt{(P ,} & \mt{c(AGS, F))}   & \text{\ttt{:-}}\ \mt{not\ not\_entails(P , c(AGS, F)).} 
				\end{array}
			\end{equation*}

			The encoding makes use of an auxiliary predicate \ttt{not\_entails} to check whether a given formula \ttt{F} is not entailed by a possibility \ttt{P1}. For formulae of the type \ttt{b(AG, F)} we require that all of the possibilities believed by \ttt{AG} entail \ttt{F}. Similarly, for formulae of the type \ttt{c(AGS, F)} (where \ttt{AGS} represents a set of agents) we require that all of the possibilities \ti{reached} by \ttt{AGS} entail \ttt{F}. A possibility \ttt{P1} reaches \ttt{P2} if it satisfies the following rules (where $\mt{contains/2}$ is defined by a set of facts):
			\begin{equation*}
				\begin{array}{l}
					\mt{reaches(P1, P2, AGS)} \text{\ttt{:-}} \mt{believes(P1, P2, AG), contains(AGS,AG).} \\
					\mt{reaches(P1, P2, AGS)} \text{\ttt{:-}} \mt{believes(P1, P3, AG), 
					contains(AGS,AG), reaches(P3, P2, AGS).}
				\end{array}
			\end{equation*}

		\paragraph{\bf Initial state generation.}\label{sec:init_state}
			The initial state is set at time $ 0 $. Since we require the initial state to be a model of a finitary \tb{S5}-theory, we assume the initial conditions to be DEL formulae of the form $(i)$--$(iv)$ (Definition~\ref{def:s5_theory}). Let us analyze how such formulae shape the initial state. Let $ \psi $ be a fluent formula, let $ \ms{f} \in D(\mc{F}) $ be a fluent and let $ \ms{i} \in D(\mc{AG}) $ be an agent. Consider a $ \mar $ statement of the form $ [\init \varphi] \in D $; we have five cases:
			\begin{enumerate}
				\item $ \varphi \equiv \ms{f} \ (\neg \ms{f}) $: \ts{f} must (not) hold in the pointed possibility.
				\item $ \varphi \equiv \C{\ } \ms{f} \ (\neg \ms{f}) $: \ts{f} must (not) hold in each possibility of the initial state.
				\item $ \varphi \equiv \C{\ } \psi $: if $ \psi $ is a fluent formula that is \ti{not} a fluent literal, then it must be entailed from each possibility of the initial state.
				\item $ \varphi \equiv \C{\ } (\B{i} \psi \vee \B{i} \neg \psi) $: there can be no two possibilities \ts{u} and \ts{v} such that $ \ms{v} \in \ms{u}(\ms{i}) $ and $ \psi $ is entailed by only one of them. Intuitively, this type of formula expresses the fact that agent \ts{i} knows whether $ \psi $ is true in the initial state.
				\item $ \varphi \equiv \C{\ } (\neg \B{i} \psi \wedge \neg \B{i} \neg \psi) $: this type of formula expresses the fact that agent \ts{i} does \ti{not} know whether $ \psi $ is true or false in the initial state. Hence, given a possibility \ts{u}, there must exist $ \ms{v} \in \ms{u}(\ms{i}) $ such that \ts{u} $\models \psi$ and \ts{v} $\not \models \psi$ (or \ts{u} $\not \models \psi$ and \ts{v} $\models \psi$).
			\end{enumerate}

			\noindent Formulae of types 1--3 are used to build the fluent sets of the possible worlds within the initial state. A fluent \ts{f} is \ti{initially known} if there exists a statement $ [\init \C{\ } (\ms{f})] $ or $ [\init \C{\ } (\neg \ms{f})] $. In the former case, all agents will know that \ts{f} is true, whereas in the latter that \ts{f} is false. If there are no such statements for \ts{f}, then it is said to be \ti{initially unknown}. Let $ uk $ be the number of initially unknown fluents: we consider $2^{uk}$ initial possible worlds, addressed by an integer index $ \mt{P} \in\{ 1, \dots, 2^{uk} \}$, one for each possible truth combination of such fluents. For each initial possibility \ttt{P} and each initially known fluent \ttt{F}, we create an atom \ttt{holds(0, 0, P, F)}, since it is common belief between all agents that \ttt{F} is true (we deal with negated fluents similarly). Moreover, through the atoms \ttt{holds} we generate all the possible truth combinations for initially unknown fluents and we assign each one of them to an initial possibility. We require all the combinations to be different, thus each initial possibility represents a unique possible world.

			An initial possibility is said to be \ti{good} if it entails all of the formulae of type 3. We create a possible world \ttt{possible\_world(0, 0, P)} for every \ti{good} initial possibility \ttt{P}. The initial pointed possibility is specified by \ttt{pointed(0, 0, PP)}, where \ttt{PP} is the (unique) \ti{good} initial possibility that entails all of the type 1 formulae. Finally, formulae of type 4 are used to filter out the edges of the initial state. Let \ttt{P1} and \ttt{P2} be two \ti{good} initial possibilities; the atom \ttt{believes(0, 0, P1, 0, 0, P2, AG)} holds if there are no initial type 4 formulae $\psi$ such that \ttt{P1} and \ttt{P2} do \ti{not} agree on $ \psi $. The construction of the initial state is achieved by filtering out the edges of a complete graph---\ie being $\mathcal{G}$ the set of \textit{good} initial possibilities, $\forall \ms{u} \in \mathcal{G}, \forall\ms{ag}\in \mathcal{AG}$ we have that $\ms{u}(\ms{ag}) = \mathcal{G}$. We can observe that type 5 formulae do not contribute to this filtering, hence we do not consider them in the initial state generation.

		\paragraph{\bf Transition function.}\label{sec:tr_function}
			The transition function calculates the resulting state after the execution of an action at time $ \mt{T} > 0 $. There are three types of actions; the implementation of executability conditions is the same for all of them. For example, suppose that at time \ttt{T} we execute the ontic action $ \ms{act} $: the statement $ [\ms{act}\ \tb{causes}\ \ms{f}\ \tb{if}\ \varphi] $ tells us that in order to apply the action effect $ \ms{f} $ on a possibility \ts{u} we first need to satisfy the condition $ \ms{u} \models \varphi $. To this end we introduced the predicate \ttt{is\_executable\_effect(T, ACT, T2, R2, P2, E)}. If such an atom holds, then it denotes that the effect \ttt{E} of the action \ttt{ACT} performed at time \ttt{T} is executable in the possibility $ (\mt{T2}, \mt{R2}, \mt{P2}) $. Without loss of generality, we represent an action instance by a unique action (using fresh actions names). Let us describe how we have modeled these actions in ASP (following the semantics described by Fabiano et al.).
			
			\subparagraph{\it Ontic actions.}
				Let \ttt{ACT} be an ontic action executed at time \ttt{T} and let $ \ms{u} = (\text{\ttt{T-1}}, \mt{RP}, \mt{PP}) $ be the pointed possibility at time \ttt{T-1}. Intuitively, when an ontic action is executed, the resulting possibility \ts{u'} is calculated by applying the action effects on \ts{u} and also on the possibilities $ \ms{w} \in \ms{u}(\ms{ag}) $, for each fully observant agent \ts{ag}; and so on, recursively. Hence, we apply the action effects to all of the possibilities \ts{w} that are \ti{reachable with a path labeled with only fully observant agents} (briefly denoted as \ti{fully observant path}). This concept is key to understand how the possible worlds are computed. Then \ttt{possible\_world} is defined as follows:
				\begin{equation*}
					\begin{array}{ll}
						\text{\ttt{possible\_world}} & \mt{(T, R2+MR+1, P2)} \text{\ttt{:-}}                                  \\
													& \text{\ttt{pointed(T-1, RP, PP),\ possible\_world(T2, R2, P2), T2<T,}} \\
													& \text{\ttt{reaches(T-1, RP, PP, T2, R2, P2, AGS),}}\ \mt{subset(AGS, F_{ACT}).}
					\end{array}
				\end{equation*}

				\noindent where $ \mt{MR} $ is the maximum value of the parameter \ti{repetition} among all the possibilities at time \ttt{T-1} and $ \mt{F_{ACT}} $ represents the set of fully observant agents of \ttt{ACT}. Hence, if $ (\mt{T2}, \mt{R2}, \mt{P2}) $ is a possibility that is reachable by a fully observant path at time \ttt{T-1}, then we create a new possibility $ (\mt{T}, \mt{R2+MR+1}, \mt{P2}) $. When the body of the rule is satisfied, we say that \ttt{P2} is \ti{updated}. For short we will refer to the updated version of \ttt{P2} as \ttt{P2}$^\prime$. The time corresponds to the step number when the possibility was created; the repetition is calculated by adding to \ttt{R2} the maximum repetition found at time \ttt{T-1}, plus one; finally, \ttt{P2} is the name of the new possibility.
			
				The pointed possibility at time \ttt{T} is \ttt{pointed(T, 2*MR+1, PP)}. Notice that, since the maximum repetition at time $ 0 $ is $ 0 $ (by construction of the initial state) and since at time \ttt{T} we set the repetition of the pointed possibility to \ttt{2*MR+1}, it follows that the maximum repetition at each time is associated to the pointed possibility itself. In this way we are able to always create a unique triple of parameters for each new possibility. At the moment, the plans that \ts{PLATO} can handle in reasonable times have lengths that limit the exponential growth of such value within an acceptable range. In fact, even for the largest instance that was tested on EFP 2.0~\cite{Fab20}, the length of the optimal plan was less than 20 (\ts{PLATO} could not find a solution for such instance before the timeout). Nonetheless, we plan a more efficient design of the update of the repetition values through hashing functions or bit maps that would limit the growth of the repetition to a polynomial rate. This would achieve a polynomial growth of the repetition value, allowing the solver to handle much longer plans.

				Next, we must state which fluents hold in the new possibilities. For each fluent \ttt{F} that is an executable effect of \ttt{ACT}, we impose $ \mt{holds(P2^\prime, F)} $ (and similarly for negative effects). The remaining fluents will hold in the updated possibility only if they did in the old one.

				Finally, we deal with the agents' beliefs. Let \ttt{P1}, \ttt{P2} be two updated possibilities and let \ttt{AG} be a fully observant agent. If \ttt{believes(P1, P2, AG)} holds, we impose \ttt{believes(P1$^\prime$, P2$^\prime$, AG)}. Otherwise, if \ttt{AG} is oblivious, we impose \ttt{believes(P1$^\prime$, P2, AG)} exploiting the already calculated possibility \ttt{P2} to reduce the number of \ttt{possible\_world} atoms.

			\subparagraph{\it Sensing/Announcement actions.}
				The behaviour of sensing and announcement actions is similar (as shown by Fabiano et al.~\cite{Fab20}). The generation of the possible worlds is also similar to that of ontic actions. Let \ttt{ACT} be a sensing or an announcement action and let \ttt{PP} and \ttt{P2} be two possibilities such that \ttt{PP} is the pointed one at time \ttt{T-1} and \ttt{P2} is reachable from \ttt{PP}. We update \ttt{P2} in the following cases:
				\begin{enumerate}
					\item $ \mt{P2} = \mt{PP} $ (here we also set \ttt{P2$^\prime$} as the pointed possibility at time \ttt{T});
					\item \ttt{P2} is reached by a fully observant path and it is consistent with the effects of \ttt{ACT};
					\item \ttt{P2} is reached by a path that starts with an edge labeled with a partially observant agent and that does \ti{not} contain oblivious agents.
				\end{enumerate}
					
				The pointed possibility must always be updated, in order for it to be consistent with the change of the agents' beliefs after the action is performed (that is, we do not want to carry old information obtained in previous states). Similarly to ontic actions, condition 2 deals with the possibilities believed by fully observant agents; if \ts{ag} is fully observant, then she must only believe those possible worlds that are consistent with the effects of \ttt{ACT}. Finally, condition 3 deals with partially observant agents: since such an agent is not aware of the action's effects, we do not impose \ttt{P2$^\prime$} to be consistent with the action's effects. Also, we restrict the first edge to be labeled by a partially observant agent in order to avoid the generation of superfluous possible worlds (namely, worlds that are not believed by any agent). In fact, the contribution to the update of the possible worlds by means of fully observant agents is entirely captured by condition 2.
		
				We create a possible world \ttt{P2$^\prime$} at time \ttt{T} for each \ttt{P2} that satisfies one of the conditions above. Since sensing and announcement actions do not alter the physical properties of the world, we impose \ttt{holds(P2$^\prime$, F)} if \ttt{holds(P2, F)}, for each fluent \ttt{F} (inertia).
			
				Let \ttt{AG} be a partially observant agent. If \ttt{believes(P1, P2, AG)} holds, then we will impose \ttt{believes(P1$^\prime$, P2$^\prime$, AG)}, since partially observant agents are not aware of the effects of the action. If \ttt{AG} is fully observant, we also add the condition that \ttt{P1} and \ttt{P2} are both (or neither) consistent with the effects of the actions. The purpose of this condition is twofold: first, we update the beliefs of the fully observant agents; second, we maintain the beliefs of partially observant agents w.r.t. the beliefs of the fully observant ones. We deal with oblivious agents exactly as for ontic actions.
		
		\paragraph{\bf Optimizations.}\label{par:optimizations}
			In order to minimize the amount of grounded \ttt{possible\_world} atoms, we designed the function so as to reuse, whenever possible, an already computed possibility. In this way, we efficiently deal with the beliefs of oblivious agents.
			
			We were also able to significantly speed up the initial state generation by imposing a complete order between the initial possible worlds w.r.t. their fluents. Specifically, let \ttt{P1} and \ttt{P2} be two initial possibilities. Let \ttt{MFi = \#max \{ F : holds(Pi, F), not holds(Pj, F) \}}, with $ \mt{i} \neq \mt{j} $. Then we impose that if $ \mt{P1} < \mt{P2} $, then it must also hold that $ \mt{MF1} < \mt{MF2} $. Since it could be the case that there exist finitely many initial states, by implementing this constraint we are able to generate a unique initial state while discarding the (possible) other equivalent ones.

		\paragraph{\bf Multi-shot encoding.}
			Following the approach of Gebser et al.~\cite{Ges17} we divided our ASP program into three main sub-programs, where the parameter \ttt{t} stands for the execution time of the actions: (i) \ttt{base}: it contains the rules for the generation of the initial state ($ \mt{t} = 0 $), alongside with the instance encoding; (ii) \ttt{step(t)}: it deals with the plan generation ($ \mt{t} > 0 $) and with the application of the transition function; and (iii) \ttt{check(t)}: it verifies whether the goal is reached at time $ \mt{t} \geq 0 $.
			
			The sub-program \ttt{check(t)} contains the external atom \ttt{query(t)} that is used in the constraint: $ \text{\ttt{:- not entails(t, R, P, F), pointed(t, R, P), goal(F), query(t)}} $. The atom \ttt{query(t)} allows the solver to activate the constraint above only at time $ \mt{t} $ (with the method \ttt{assign\_external}) and to deactivate it when we move to time $ \mt{t+1} $ (method \ttt{release\_external}). Using the Python script provided by Gebser et al., we first ground and solve the sub-program \ttt{base} and we check if the goal is reached in the initial state ($ \mt{t} = 0$); in the following iterations, the sub-programs \ttt{step(t)} ($ \mt{t} > 0 $) are grounded and solved; we check the goal constraint until the condition is satisfied.


	\section{Experimental Evaluation}\label{sec:evaluation}
		In this Section we compare \ts{PLATO} to the multi-agent epistemic planner EFP 2.0 presented in previous work~\cite{Fab20}. All the experiments were performed on a 3.60GHz Intel Core i7-4790 machine with 32 GB of memory and with Ubuntu 18.04.3 LTS, imposing a time out (\ttt{TO}) of 25 minutes and exploiting ASP's parallelism on multiple threads. All the results are given in seconds. From now on, to avoid unnecessary clutter, we will make use of the following notations:
		\begin{itemize}
			\item $ L $: the length of the optimal plan;
			\item $ d $: the upper bound to the depth of nested modal operators $ \B{} $ in the DEL formulae;
			\item K-BIS/P-MAR: the solver EFP 2.0 using the best configuration based on Kripke structures and possibilities, respectively;
			\item \texttt{single}/\texttt{multi}: \ts{PLATO} using the single-shot/multi-shot encoding, respectively;
			\item \texttt{many}/\texttt{frumpy}: \texttt{multi} using the \ti{clingo}'s configuration \ti{many}/\ti{frumpy}, respectively;
			\item \texttt{bis}: \texttt{multi} implemented with a visited state check based on bisimulation (following the implementation of Dovier~\cite{DBLP:conf/iclp/Dovier15}).
		\end{itemize}
		
		\begin{table}
			\centering
			\hfill%
			\subfloat[Runtimes for Selective Communication.]{\includegraphics{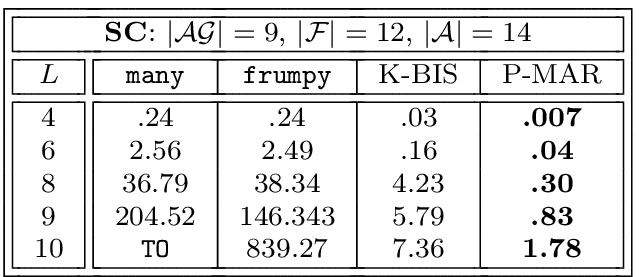}\label{tab:sc}}
			\hfill%
			\subfloat[Runtimes for Grapevine.]{\includegraphics{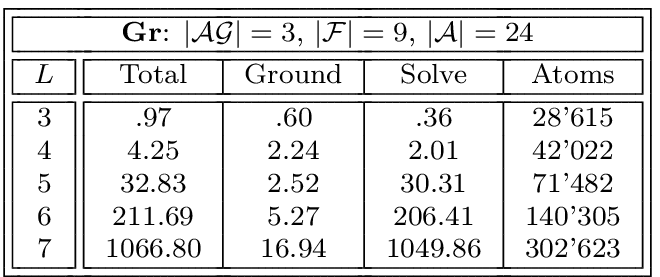}\label{tab:gr}}
			\hfill%
			\hfill%
			\\
			\hfill%
			\subfloat[Runtimes for Coin in the Box.]{\includegraphics {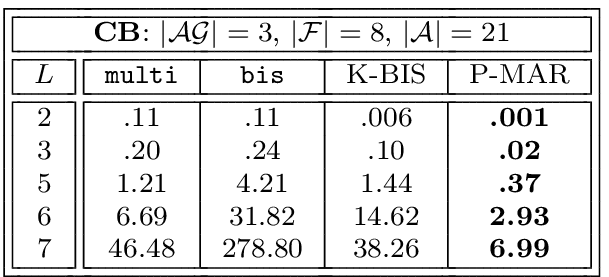}\label{tab:cb}}
			\hfill%
			\subfloat[Runtimes for Assembly Line.]{\includegraphics{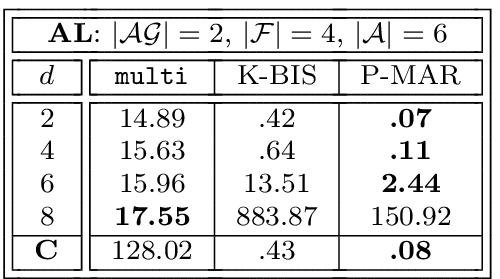}\label{tab:al}}
			\hfill%
			\hfill%
			\hfill%
			\\
			\subfloat[Runtimes for Collaboration and Communication.]{\includegraphics{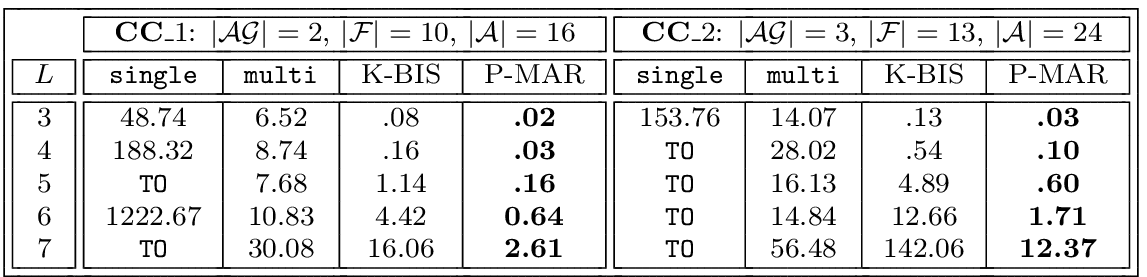}\label{tab:cc}}
			\caption{(a) Comparison of \texttt{frumpy}, \texttt{many} and EFP 2.0 on \tb{SC}. (b) Total, grounding and solving times for \tb{Gr} using \texttt{multi}. The last column reports the number of grounded atoms. (c) Comparison of \texttt{multi} and \texttt{bis} on \tb{CB}. (d) Comparison of \ts{PLATO} and EFP 2.0 on \tb{AL} ($ \C{} $ identifies that the executability conditions are expressed through common believes). (e) Comparison of \texttt{single}, \texttt{multi} and EFP 2.0 on \tb{CC}.}
		\end{table}
	
		We considered various domains collected from the literature~\cite{kominis2015beliefs,huang2017general}, such as \ti{Collaboration and Communication} (\tb{CC}), \ti{Selective Communication} (\tb{SC}), \ti{Grapevine} (\tb{Gr}), \ti{Assembly Line} (\tb{AL}), and \ti{Coin in the Box} (\tb{CB}). The full description of these domains is reported in the supplementary documents and it can also be found in previous work by Fabiano et al.~\cite{Fab20}.
	
		We report only the results of the \ti{clingo}'s search heuristic configurations \ti{many} and \ti{frumpy} as they were the most performing ones in our set of benchmarks. Although generally they show a similar behaviour, as shown in Table~\ref{tab:sc}, in larger instances the time results differ substantially. In the results, when only \ttt{multi} is specified, we considered the most efficient configuration on the specific domain.
	
		To evaluate the behaviour of \ts{PLATO} w.r.t. the entailment of DEL formulae, we exploited the \tb{AL} domain (Table~\ref{tab:al}), where the executability conditions of the actions have depth $ d $. The entailment of belief formulae with higher depth is handled efficiently by \ts{PLATO}, although the use of common believes in the executability conditions leads to worst results. This is due to the fact that the number of \ttt{reached} atoms is substantially higher than the number of \ttt{believes} atoms (required in the entailment of $ \C{} $ and $ \B{} $ formulae, respectively). Notice that in ASP the entailment of each formula, independently from its depth, is handled by a grounded atom and, therefore, a higher depth does not impact the solving process. On the other hand, the entailment in EFP 2.0 is handled by exploring all the paths of length $ d $ of the state, causing higher cost performances during each entailment check.
	
		To investigate the contribution of the grounding and solving phases, we summed the computation times of the \ti{clingo} functions \ttt{ground()} and \ttt{solve()} for each iteration. Table~\ref{tab:gr} shows a major contribution of the solving phase, hence indicating an efficient grounding. This permitted us to consider larger instances and to compete with other imperative approaches. The implementation of bisimulation within the multi-shot encoding leads to less efficient results (as shown in Table~\ref{tab:cb}), due to a much heavier contribution of the grounding phase.
	
		Finally, we compare the single-shot/multi-shot encodings in Table~\ref{tab:cc}. The latter approach leads to significantly better results: the \ti{clingo}’s option \ttt{--stat} revealed a smaller number of conflicts in the majority of the benchmarks. As explained by Gebser et al.~\cite{Ges17}, this is due to the reuse of \ti{nogoods} learnt from previous solving steps.


	\section{Correctness of \ts{PLATO} w.r.t. $\mar$}\label{sec:correctness}
		Declarative languages such as ASP allow a high-level implementation, facilitating the derivation of a formal verification of correctness of the planner. Consider a domain $ D $; we denote the set of the belief formulae that can be built using the fluents in $ D(\mc{F}) $ and the propositional/modal operators by $ D(\mc{BF}) $. We denote the transition function of \ts{PLATO} by $\Gamma: D(\mc{AI}) \times D(\mc{S}) \rightarrow D(\mc{S})\ \cup\ \{\emptyset\}$ (where $D(\mc{AI})$ and $D(\mc{S})$ are defined as in Section~\ref{sec:efp}). Finally, we express the entailment w.r.t. $ \mar $ and \ts{PLATO} with $ \models_{\Phi} $ and $ \models_{\Gamma} $, respectively. Each main component of the planner is addressed by a relative Proposition.
			
		\begin{proposition}[\ts{PLATO} entailment correctness w.r.t. $ \mar $]\label{prop:entail}
			Given a possibility $ \ms{u} \in D(\mc{S}) $ we have that $ \forall \psi \in D(\mc{BF})\ \ms{u} \models_{\Phi} \psi \text{ iff } \ms{u} \models_{\Gamma} \psi\ $.
		\end{proposition}
		
		\begin{proposition}[\ts{PLATO} initial state construction correctness w.r.t. $ \mar $]\label{prop:ini}
			Given two possibilities $ \ms{u}, \ms{v} \in D(\mc{S}) $ such that $ \ms{u} $ is the initial state in $ \mar $ and $ \ms{v} $ is the initial state in \ts{PLATO} then $ \forall \psi \in D(\mc{BF})\ \ms{u} \models_{\Phi} \psi \text{ iff } \ms{v} \models_{\Gamma} \psi $.
		\end{proposition}

		\begin{proposition}[\ts{PLATO} actions correctness w.r.t. $ \mar $]\label{prop:ontic_corr}
			Given an action instance $ \ms{a} \in D(\mc{AI}) $ and two possibilities $ \ms{u}, \ms{v} \in D(\mc{S}) $ such that $ \forall \psi \in D(\mc{BF})\ \ms{u} \models_{\Phi} \psi \text{ iff } \ms{v} \models_{\Gamma} \psi $ then $ \forall \psi \in D(\mc{BF})\ \Phi(\ms{a}, \ms{u}) \models_{\Phi} \psi \text{ iff } \Gamma(\ms{a}, \ms{v}) \models_{\Gamma} \psi $.
		\end{proposition}
			
		\noindent The complete proofs are provided in the supplementary documents that are available at \url{http://clp.dimi.uniud.it/sw/}. This results allowed us to verify the empirical correctness of the planner EFP 2.0. In all of the conducted tests, the two planners exhibited the same behaviour. In the same way, \ts{PLATO} can be used to verify empirically the correctness of any multi-agent epistemic planner that is based on a semantics equivalent to the one of $ \mar $. Finally, as the \ti{plan existence problem} in the MEP setting is \ti{undecidable}~\cite{bolander2011epistemic}, all the planners that reason on DEL are \ti{incomplete}. Since infinitely many e-states could be potentially generated during a planning process, in general both EFP 2.0 and \ts{PLATO} are unable to determine if a solution for a planning problem exists (even when checking for already visited states).


	\section{Conclusions and Future Works}\label{sec:conclusions}
		In this paper we presented a multi-agent epistemic planner implemented in ASP. The implementation of MEP in a declarative language involves various advantages. First, the reduced size of the program allows a better readability of the code, allowing a much easier approach to MEP. Second, code maintainability is simpler and, third, modifications on the semantics of $ \mar $ can be manageably implemented. Specifically, if new operators or actions types are added to $ \mar $ (\eg concepts such as: \ti{trust}, \ti{lies} or \ti{inconsistent beliefs}), it would suffice to add or modify a small number of rules. Ultimately, the extent of the code adaptation would be significantly lower w.r.t. the imperative approaches.
		
		We were able to exploit several ASP features, such as the multi-thread parallelisation and the different solving configurations. Approaching MEP through declarative programming will also allow automated epistemic reasoning to benefit from the constant enhancement of ASP solvers' efficiency. These features, together with an efficient grounding, allowed us to achieve comparable results w.r.t. EFP 2.0. ASP also allows to find \ti{all} the solutions of a planning instance without any addition to the code. The formal verification of the correctness of \ts{PLATO} (Section~\ref{sec:correctness}) permitted us to empirically verify the correctness of EFP 2.0 by comparing the obtained plans on both solvers.
		
		As future works, we plan to improve \ts{PLATO} in several ways. First, we plan to enhance the entailment, by defining different entailment rules for different formulae types (\eg, executability conditions, actions effect conditions, etc.). This will impact on both grounding and solving efficiency. We also plan a more efficient design of the update of the repetition values through hashing functions or bit maps; this would limit the growth of the repetition to a polynomial rate.
		
		Second, we plan to implement some heuristics, such as choosing to perform the action that leads to the satisfaction of the higher number of goal conditions, so as to improve the computational results. Finally, we plan to exploit \ts{PLATO} to formally prove that $ \mal $ and $ \mar $ are semantically equivalent. This would provide a much stronger result w.r.t. the one proved by Fabiano et al. in previous work~\cite{Fab20}.

	\paragraph{Acknowledgments.}
		The authors wish to thank Martin Gebser and Roland Kaminski for their suggestions on the ASP encoding and the anonymous Reviewers for their comments that allowed to improve the presentation. The research is partially supported by Indam GNCS, by Uniud project ENCASE, and by NSF grants 1914635, 1833630, and 1345232.

\end{document}